\documentclass[journal]{IEEEtran}

\usepackage{booktabs}
\usepackage{graphicx}
\usepackage{subcaption}
\usepackage{graphicx}
\usepackage{textcomp}
\usepackage{xcolor}
\usepackage{cite}
\usepackage{amsmath,amssymb,amsfonts}
\usepackage{algorithm}
\usepackage{algorithmic}
\usepackage{makecell}
\usepackage{float}
\usepackage[colorlinks=true, urlcolor=blue]{hyperref}

\begin{document}
%
\title{External Data-Enhanced Meta-Representation for Adaptive Probabilistic Load Forecasting}
%
%
%

\author{Haoran~Li,~\IEEEmembership{Member,~IEEE,}
Muhao~Guo,~\IEEEmembership{Student Member,~IEEE,}
Marija~Ilic,~\IEEEmembership{Life Fellow,~IEEE,}
Yang~Weng,~\IEEEmembership{Senior Member,~IEEE,}
Guangchun~Ruan,~\IEEEmembership{Member,~IEEE,}
\thanks{Haoran Li, Muhao Guo, and Yang Weng are with the Department of Electrical, Computer and Energy Engineering, Arizona State University, Tempe, AZ, 85281, USA. E-mail: \mbox{\{lhaoran,mguo26,yang.weng\}@asu.edu}. 

Marija Ilic and Guangchun Ruan are with the Laboratory for Information and Decision Systems, Massachusetts Institute of Technology, Cambridge, MA, 02139, USA. E-mail: \mbox{\{ilic, gruan\}@mit.edu}.

}
\vspace{-10mm}
}

\maketitle

\begin{abstract}
Accurate residential load forecasting is critical for power system reliability with rising renewable integration and demand-side flexibility. However, most statistical and machine learning models treat external factors, such as weather, calendar effects, and pricing, as extra input, ignoring their heterogeneity, and thus limiting the extraction of useful external information. We propose a paradigm shift: external data should serve as meta-knowledge to dynamically adapt the forecasting model itself. Based on this idea, we design a meta-representation framework using hypernetworks that modulate selected parameters of a base Deep Learning (DL) model in response to external conditions. This provides both expressivity and adaptability. 
We further integrate a Mixture-of-Experts (MoE) mechanism to enhance efficiency through selective expert activation, while improving robustness by filtering redundant external inputs. 
The resulting model, dubbed as a Meta Mixture of Experts for External data (\text{M\textsuperscript{2}oE\textsuperscript{2}}), achieves substantial improvements in accuracy and robustness with limited additional overhead, outperforming existing state-of-the-art methods in diverse load datasets. The dataset and source code are publicly available at \href{https://github.com/haorandd/M2oE2_load_forecast.git} {\textcolor{blue}{https://github.com/haorandd/M2oE2\_load\_forecast.git}}.

\end{abstract}


%
\IEEEpeerreviewmaketitle

\section{Introduction}


Load forecasting is a long-standing and fundamental task in power systems, underpinning key operations such as economic dispatch~\cite{han2021task}, demand response~\cite{pramono2019deep}, grid stability and reliability analysis~\cite{saxena2024intelligent}, event detection~\cite{ref:Haoran2022D,ref:Haoran2022T,ref:Haoran2023S}, and infrastructure planning~\cite{zhang2023novel}. With increasing smart meters, there is a growing shift toward bottom-up residential load forecasting~\cite{ref:2019ChengjinD}. This is because accurate forecasts at the residential level support distribution-level operations, e.g., the coordination of Distributed Energy Resources (DERs), local market participation, and demand-side management~\cite{sajjad2020novel}. Additionally, load forecast also enhances system-wide reliability and efficiency by improving bulk market predictions and capacity planning~\cite{ref:Ponoćko2018F}. 


Despite its importance, highly granular load forecasting remains challenging due to high volatility and non-stationarity. Unlike aggregated loads, residential usage patterns are heavily influenced by diverse factors such as occupant behavior, appliance schedules, weather, and the growing integration of rooftop photovoltaics and electric vehicles, all of which contribute to complex and irregular temporal dynamics~\cite{kong2017short}. 
To gain accurate forecasts, a promising approach is to effectively incorporate heterogeneous external information, such as weather conditions, calendar features, and demand response (DR) signals~\cite{wang2018review}. However, these external factors often vary in data type (continuous vs. categorical), exhibit weak or delayed correlations with the load, and require specialized treatment for meaningful integration into forecasting models.

\begin{figure}
    \centering \includegraphics[width=1\linewidth]{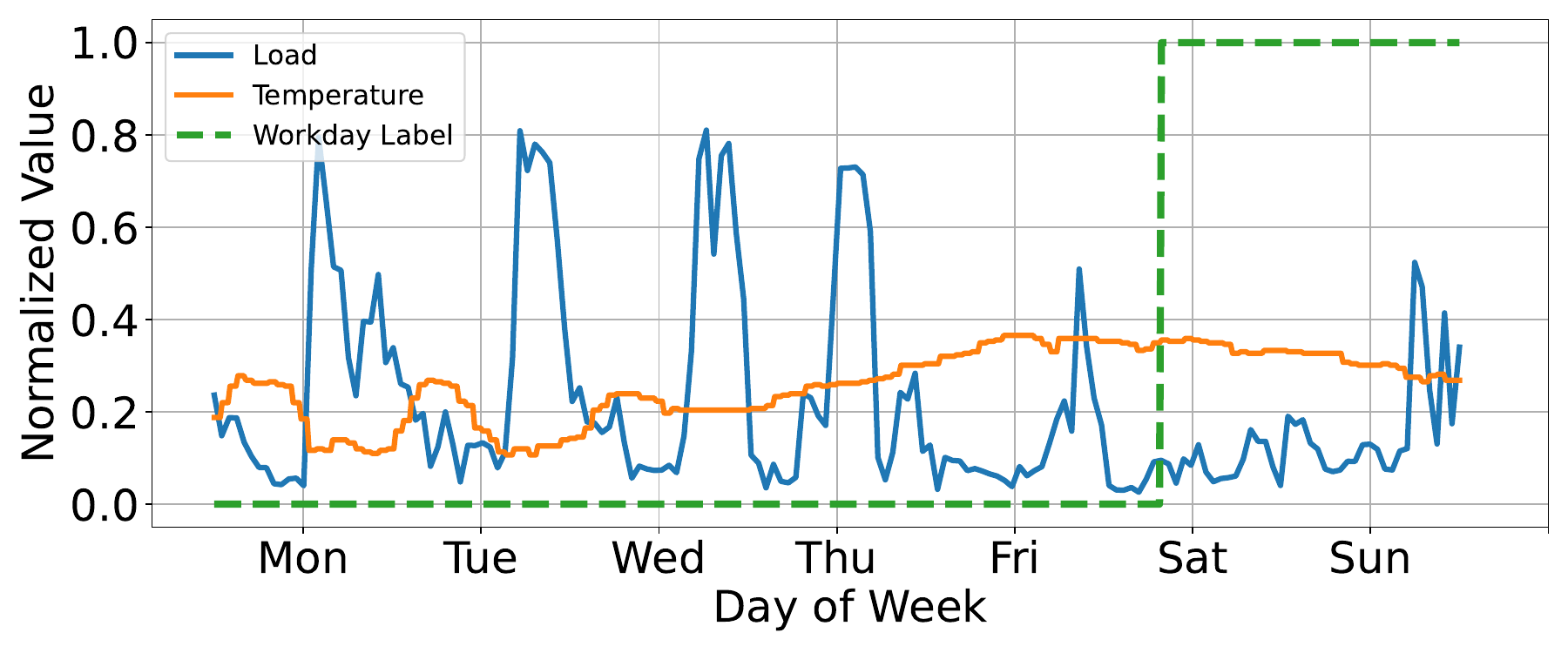}
    \caption{The visualization of normalized external data and loads. The blue curve is the load data, the orange curve is the temperature data, and the green curve is the workday labels (0: weekday, 1: weekend). External data are relatively flat compared to the highly variable load curves.}
    \label{fig:one_week_data}
    \vspace{-5mm}
\end{figure}


Therefore, a large body of work aims to improve residential load forecasting through increasingly sophisticated modeling techniques. Nevertheless, most of them share a common design flaw: treating all available data, including external information, as static inputs to a single fixed model. Such a design overlooks the contextual role that external factors can play in dynamically modulating load behavior. For example, as shown in Fig. \ref{fig:one_week_data}, we perform normalization to make data from different sources comparable. However, external data are relatively flat compared to highly variable load curves. Therefore, in terms of extracting the temporal trends and patterns, a forecaster will pay great attention to historical load data and nearly ignore external information. In addition, given diverse external sources, it remains challenging to simultaneously process them. The model may suffer from overfitting issues when incorporating too many external datasets with high dimensionality and redundancy.

These limitations have been observed in existing approaches, which can be categorized into three groups. The first employs classical statistical or Machine Learning (ML) techniques, such as ARIMA, SARIMA, exponential smoothing~\cite{christiaanse1971short}, Gaussian process \cite{weng2015probabilistic,xie2017adaptive,xie2018integrated,weng2018probabilistic,xie2020input}, and kernel regressor~\cite{abbasi2019short}, which rely on handcrafted features and struggle with high-dimensional, nonlinear load dynamics~\cite{lu2024short}. These models often treat contextual information as fixed covariates without modeling how they influence the forecasting function itself. 

The second group leverages Deep Learning (DL) models, e.g., Recurrent Neural Networks (RNNs) \cite{che2018recurrent}, Long Short-Term Memory (LSTM)~\cite{kong2017short,xie2021enhance}, and Transformers~\cite{l2022transformer}, to automatically extract hierarchical features from smart meter data, often outperforming classical methods by capturing complex temporal dependencies. Hybrid architectures like Convolutional Neural Network (CNN-)LSTM models~\cite{dong2022ultra} further improve representation learning by combining local and sequential patterns, but still treat external signals as additive inputs \cite{cao2025short,zeng2021uncovering}. The third group introduces physics-informed ML models, which encode physical constraints related to building systems~\cite{chen2023physics}, grid dynamics~\cite{li2022physics}, or battery behaviors~\cite{xu2022physics}. These methods are less applicable to residential loads where physical equations may be unavailable. 


In this paper, we propose to treat external information as \emph{meta-knowledge} that dynamically changes the forecasting model, rather than modeling it as an additional input. Specifically, we employ \emph{hypernetworks}~\cite{ha2016hypernetworks} to modulate a compact subset of parameters in the base network based on contextual signals. This meta-representation allows the model to adjust its behavior in response to evolving external conditions, such as weather, calendar effects, and occupant behaviors. To ensure numerical stability and balanced influence across heterogeneous data sources, we apply Layer Normalization~\cite{xiong2020layer} to each meta-representation. Moreover, we carefully constrain the parameter modulation to a low-dimensional, high-impact subset of weights to maintain computational efficiency, which is a significant extension of our previous work \cite{li2025exarnn}.

In addition to the high model adaptation and moderate efficiency, our architecture naturally integrates enhancements such as robustness and plug-and-play properties without requiring major architectural changes. To improve \emph{robustness}, we introduce a Mixture-of-Experts (MoE) structure~\cite{zhou2022mixture}, where each expert corresponds to a distinct external source. A learned gating mechanism dynamically selects a sparse subset of active experts, enabling the model to suppress irrelevant or redundant signals and focus on the informative context. To ensure that our meta-framework always enhances the base forecaster, we adopt \emph{residual parameter modulation}. The mechanism guarantees that external information is always employed to reduce the residual errors of a base model. Consequently, the model's accuracy keeps increasing as long as we can introduce a more accurate base forecaster. While some related frameworks emerged in $2025$, they typically integrate MoE or residual ideas in an ad-hoc fashion \cite{tsoumplekas2025few,huang2025metaeformer}. In contrast, our architecture was designed to support modularity and scalability in modern load forecasting environments.


The overall model is dubbed as a Meta Mixture of Experts for External data (\text{M\textsuperscript{2}oE\textsuperscript{2}}). We conduct extensive experiments on multiple real-world residential load datasets to evaluate the effectiveness of \text{M\textsuperscript{2}oE\textsuperscript{2}}. These datasets span different geographic regions, temporal resolutions, and weather patterns, ensuring a diverse and realistic testbed. We benchmark our approach against state-of-the-art baselines, including deep learning models with direct external input fusion, attention-based methods, etc. Results demonstrate that \text{M\textsuperscript{2}oE\textsuperscript{2}} consistently outperforms alternatives in terms of point forecasting accuracy, probabilistic sharpness, and calibration. These empirical findings substantiate our central thesis: that treating external information as meta-knowledge, rather than static inputs, yields more accurate, efficient, and resilient forecasts in complex and dynamic environments.

The rest of the paper is as follows. Section \ref{sec:prob} formulates the problem.
Section \ref{sec:pre} provides preliminaries of related DL techniques. Section \ref{sec:model} illustrates our \text{M\textsuperscript{2}oE\textsuperscript{2}} framework. Section \ref{sec:exp} presents numerical results, and Section \ref{sec:con} concludes the paper.

\vspace{-3mm}
\section{Problem Formulation}
\label{sec:prob}

We define the probabilistic load forecasting problem with external data in the following. 

\begin{itemize}
\item Goal: Fuse external data and historical load data to forecast loads and quantify the uncertainty. 
\item Given: Load measurements $\{\boldsymbol{x}_i\}_{i=1}^N$ and environmental measurements $\{w_{ji}|j=1,\cdots,M;i=1,\cdots,N\}$, where $N$ is the number of measurements and $M$ is the number of external datasets. The bold $\boldsymbol{x}_i$ is a vector that implies a set of loads collected from different locations.   

\item Find: A well-trained probabilistic model to utilize historical load and external data to predict short-term future load distributions. 
\end{itemize}


To achieve accurate probabilistic forecasting, we demand a mechanism to understand the contextual information in $w_{ij}$ to intelligently adapt the forecast model. Hence, we present the following DL concepts that lay solid foundations for our models.

\section{Preliminary}
\label{sec:pre}

\subsection{Sequence DL Model to Extract Temporal Patterns}

A sequence DL model refers to a class of DL architectures that are designed to handle sequential data. In particular, the family of RNNs has proven to be effective in load forecasting \cite{kong2017short}. Let $f_{\Theta}(\cdot)$ denote an RNN model parameterized by $\Theta$. The crucial design of an RNN is to employ a hidden state vector $\boldsymbol{h}_i$ to recurrently store the past information and recognize the temporal patterns, which can be conveniently used to predict future loads. Mathematically, let $\boldsymbol{h}_i$ denote the $i^{th}$ hidden state, and the recurrent updating rule of the hidden state can be written as
\begin{equation}
\begin{aligned}
\label{eqn:rnn}
\boldsymbol{h}_i&=\rho\big(W_{hx}\boldsymbol{x}_i+W_{hh}\boldsymbol{h}_{i-1}+\boldsymbol{b}_1\big),
\end{aligned}
\end{equation}
where $\rho$ is the activation function, $W_{hx}$, $W_{hh}$, and $\boldsymbol{b}_1$ are weight matrices and bias terms, respectively. $\boldsymbol{h}_i$ is a compact and informative feature vector to represent critical information in $\{\boldsymbol{x}_j\}_{j=1}^i$. One can further build a mapping to convert from $\boldsymbol{h}_i$ to future loads, whose formula depends on the specific forecasting tasks.

\noindent \textbf{Remark}: The basic RNN model can be replaced with other sequence models like LSTM and CNN-RNN. In general, they modify Eq. \eqref{eqn:rnn} to achieve better feature extractions. LSTM includes more sophisticated gating mechanisms to structure Eq. \eqref{eqn:rnn}, and CNN-RNN leverages temporal convolution to process $\boldsymbol{x}_i$ and extract local patterns.

\subsection{DL-based Deterministic and Probabilistic Forecasting}
\label{subsec:dl_forecast}
For deterministic forecasting, we can develop another nonlinear layer in the RNN to estimate the future loads:
\begin{equation}
\label{eqn:deter_forecast}
\hat{\boldsymbol{x}}_{i+1} = \rho(W_{xh}\boldsymbol{h}_i+\boldsymbol{b}_2),
\end{equation}
where $W_{xh}$ and $\boldsymbol{b}_2$ are the weight matrix and the bias term, respectively. Eq. \eqref{eqn:deter_forecast} aims for a one-step short-term forecast, but we can extend it to a $K$-step forecast by outputting a concatenated vector $[\hat{\boldsymbol{x}}_{i+1},\cdots,\hat{\boldsymbol{x}}_{i+K}]$. In this scenario, the RNN parameter set $\Theta=\{W_{hx},W_{hh},\boldsymbol{b}_1,W_{xh},\boldsymbol{b}_2\}$. Then, the training objective is to minimize the Mean Square Error (MSE) with respect to $\Theta$:
\begin{equation}
\label{eqn:rnn_mse}
L_{\text{MSE}}(\Theta) = \mathbb{E}\big[||\hat{\boldsymbol{x}}_i-\boldsymbol{x}_i||_2^2\big].
\end{equation}

For probabilistic forecasting, the prediction and the loss function rely on the assumptions of the underlying load distributions. Specifically, if we assume a Gaussian distribution, we can construct a mean- and a variance-layer such that
\begin{equation}
\label{eqn:prob_forecast1}
\hat{\boldsymbol{\mu}}_{i+1} = \rho(W_{\mu_x h} \boldsymbol{h}_i + \boldsymbol{b}_{\mu x}), \quad
\log \hat{\boldsymbol{\sigma}}^2_{i+1} = \rho(W_{\sigma_x h} \boldsymbol{h}_i + \boldsymbol{b}_{\sigma x}),
\end{equation}
where the predicted mean and variance define the Gaussian distribution $\mathcal{N}(\hat{\boldsymbol{\mu}}_{i+1}, \text{diag}(\hat{\boldsymbol{\sigma}}^2_{i+1}))$. The log operation in Eq. \eqref{eqn:prob_forecast1} ensures the positivity of the variance and increases the numerical stability \cite{kendall2017uncertainties}.   
In this scenario, we denote the RNN parameter set $\Theta=\{W_{hx},W_{hh},\boldsymbol{b}_1,W_{\mu_x h},\boldsymbol{b}_{\mu x},W_{\sigma_x h},\boldsymbol{b}_{\sigma x}\}$. Then, the training objective is to minimize the Negative Log-Likelihood (NLL) with respect to $\Theta$:
\begin{equation}
\label{eqn:rnn_nll}
L_{\text{NLL}}(\Theta) = \frac{1}{N-1} \sum_{i=2}^{N} \left[ \log \hat{\boldsymbol{\sigma}}^2_i + \frac{(\boldsymbol{x}_i - \hat{\boldsymbol{\mu}}_i)^2}{\hat{\boldsymbol{\sigma}}^2_i} \right].
\end{equation}

However, the assumption of Gaussian distribution is often unrealistic, as real-world load data can exhibit multimodality, asymmetry, and non-stationary behaviors due to complex and uncertain external factors. Hence, a more general approach is to introduce a latent variable to model the underlying stochasticity in the forecasting process. Specifically, we assume the \textbf{random latent vector} $\boldsymbol{z}$ captures hidden factors influencing future load given the past context. Following the variational inference framework \cite{rezende2014stochastic}, we introduce an approximate posterior over the latent variable conditioned on both the current hidden state and the target future value:
\begin{equation}
\begin{aligned}
\label{eqn:rnn_encode}
q(\boldsymbol{z}_{i+1} \mid \boldsymbol{x}_{i},\boldsymbol{h}_{i-1}) &= \mathcal{N}\bigg(\boldsymbol{\mu}_z(\boldsymbol{x}_{i},\boldsymbol{h}_{i-1}), \operatorname{diag}\big(\boldsymbol{\sigma}_z^2(\boldsymbol{x}_{i},\boldsymbol{h}_{i-1})\big)\bigg),\\
\boldsymbol{\mu}_z(\boldsymbol{x}_{i},\boldsymbol{h}_{i-1}) &= \rho(W_{\mu_z x}\boldsymbol{x}_{i}+W_{\mu_z h} \boldsymbol{h}_{i-1}+\boldsymbol{b}_{\mu z}), \\
\quad
\log \boldsymbol{\sigma}^2_z(\boldsymbol{x}_{i},\boldsymbol{h}_{i-1}) &= \rho( W_{\sigma_z x}\boldsymbol{x}_{i} + W_{\sigma_z h} \boldsymbol{h}_{i-1} + \boldsymbol{b}_{\sigma z}).
\end{aligned}
\end{equation}

To allow gradient-based optimization, we apply the reparameterization trick:
\begin{equation}
\label{eqn:reparameterization}
\boldsymbol{z}_{i+1} = \boldsymbol{\mu}_z(\boldsymbol{x}_{i},\boldsymbol{h}_{i-1}) + \boldsymbol{\sigma}_z(\boldsymbol{x}_{i},\boldsymbol{h}_{i-1}) \odot \boldsymbol{\epsilon}, \quad \boldsymbol{\epsilon} \sim \mathcal{N}(\boldsymbol{0}, \boldsymbol{I}).
\end{equation}

In general, we employ a sequence \textbf{encoder}, including an RNN in Eq. \eqref{eqn:rnn} and output layers in Eq. \eqref{eqn:rnn_encode} to compress historical measurements and produce a latent vector $\boldsymbol{z}_{i+1}$. Correspondingly, we need a \textbf{decoder} to convert from forecast future loads. The decoder models the conditional distribution of the future load given the latent variable:
\begin{equation}
\begin{aligned}
\label{eqn:rnn_decode}
p(\boldsymbol{x}_{i+1} \mid \boldsymbol{z}_{i+1}) &= \mathcal{N}\bigg(\boldsymbol{\mu}_x(\boldsymbol{z}_{i+1}), \operatorname{diag}\big(\boldsymbol{\sigma}_x^2(\boldsymbol{z}_{i+1})\big)\bigg),\\
\boldsymbol{\mu}_x(\boldsymbol{z}_{i+1}) &= \rho(W_{\mu_x z} \boldsymbol{z}_{i+1} + \boldsymbol{b}_{\mu x}),\\
\log \boldsymbol{\sigma}_x^2(\boldsymbol{z}_{i+1}) &= \rho(W_{\sigma_x z} \boldsymbol{z}_{i+1} + \boldsymbol{b}_{\sigma x}).
\end{aligned}
\end{equation}
 
Although the decoder assumes a Gaussian distribution conditioned on $\boldsymbol{z}_{i+1}$ and $\boldsymbol{h}_i$, the marginal distribution of $\boldsymbol{x}_{i+1}$ becomes non-Gaussian due to the stochasticity of $\boldsymbol{z}_{i+1}$. This allows the model to capture complex uncertainty, including multimodal or heavy-tailed behaviors.
At test time, we sample the latent variable from the learned approximate posterior $q(\boldsymbol{z}_{i+1} \mid \boldsymbol{x}_i,\boldsymbol{h}_{i-1})$ in Eq. \eqref{eqn:rnn_encode}. The forecasting procedure is therefore
\begin{equation}
\label{eqn:prob_forecast2}
\boldsymbol{z}_{i+1} \sim q(\boldsymbol{z}_{i+1} \mid \boldsymbol{x}_i,\boldsymbol{h}_{i-1}), \quad \hat{\boldsymbol{x}}_{i+1} \sim p(\boldsymbol{x}_{i+1} \mid \boldsymbol{z}_{i+1}).
\end{equation}

To characterize the predictive distribution of $\boldsymbol{x}_{i+1}$, we employ a Monte Carlo approximation. Specifically, we draw $J$ samples of the latent variable $\{\boldsymbol{z}_{i+1}^{(j)}\}_{j=1}^J \sim q(\boldsymbol{z}_{i+1} \mid \boldsymbol{x}_i,\boldsymbol{h}_{i-1})$ and generate corresponding outputs $\{\hat{\boldsymbol{x}}_{i+1}^{(j)}\}_{j=1}^J \sim p(\boldsymbol{x}_{i+1} \mid \boldsymbol{z}_{i+1}^{(j)})$ from the decoder. The predictive mean and variance of $\boldsymbol{x}_{i+1}$ can then be approximated as:

\begin{equation}
\begin{aligned}
\label{eqn:mc_mean_var}
\hat{\boldsymbol{\mu}}_{i+1} &\approx \frac{1}{J} \sum_{j=1}^{J} \hat{\boldsymbol{x}}_{i+1}^{(j)},\\
\hat{\boldsymbol{\sigma}}^2_{i+1} &\approx \frac{1}{J-1} \sum_{j=1}^{J} \left(\hat{\boldsymbol{x}}_{i+1}^{(j)} - \hat{\boldsymbol{\mu}}_{i+1}\right)^2.
\end{aligned}
\end{equation}

The estimated mean and variance in Eq. \eqref{eqn:mc_mean_var} characterize the central tendency and dispersion of the predictions, which are valuable for uncertainty quantification and downstream decision-making. Moreover, Monte Carlo samples can be used to estimate other statistics, such as quantiles, to more comprehensively describe the predicted distribution.

Subsequently, we define the parameter set $\Theta$ as follows: $\Theta = \Theta_{\text{enc}} \cup \Theta_{\text{dec}},$ where $\Theta_{\text{enc}} = \{W_{hx}, W_{hh}, \boldsymbol{b}_1, W_{\mu_z h}, W_{\mu_z x}, \boldsymbol{b}_{\mu z}, W_{\sigma_z h}, W_{\sigma_z x}, \boldsymbol{b}_{\sigma z}\}$ 
represents the parameter set of the encoder, and 
$\Theta_{\text{dec}} = \{W_{\mu_x z}, W_{\mu_x h}, \boldsymbol{b}_{\mu x}, W_{\sigma_x z}, W_{\sigma_x h}, \boldsymbol{b}_{\sigma x}\}$ 
denotes the decoder parameters. The training objective is to minimize the negative Evidence Lower Bound (ELBO):

\begin{equation}
\begin{aligned}
\label{eqn:rnn_elbo}
L_{\text{ELBO}}(\Theta) = &-\sum_{i=1} \mathbb{E}_{q(\boldsymbol{z}_{i+1} \mid \boldsymbol{x}_i,\boldsymbol{h}_{i-1})} \big[ \log p(\boldsymbol{x}_{i+1} \mid \boldsymbol{z}_{i+1}) \big] \\
&+ \lambda\cdot\text{KL}\big(q(\boldsymbol{z}_{i+1} \mid \boldsymbol{x}_i,\boldsymbol{h}_{i-1}) \| p(\boldsymbol{z}_{i+1})\big),
\end{aligned}
\end{equation}
where the first term is the NLL in Eq. \eqref{eqn:rnn_nll}, and the second term is the Kullback–Leibler (KL) divergence between the posterior of the latent distribution $q(\boldsymbol{z}_{i+1} \mid \boldsymbol{h}_i, \boldsymbol{x}_{i+1})$ and the prior standard Gaussian distribution $p(\boldsymbol{z}_{i+1}) = \mathcal{N}(\boldsymbol{0}, \boldsymbol{I})$, which regularizes the latent space and avoids overfitting. $\lambda$ is a positive weight.


\begin{table*}[t]
    \centering
    \caption{Comparisons of the size and impacts of different DL sub-parameters.}
    \resizebox{\textwidth}{!}{
    \begin{tabular}{l|c|p{0.6\textwidth}}
        \toprule
        \textbf{Parameter} & \textbf{Size} & \textbf{Impact} \\
        \midrule
        Input-to-hidden matrix $W_{hx}$ in Eq. \eqref{eqn:rnn} & $N \cdot M\cdot d_x \cdot d_h$ & External conditions guide the DL model to store input load data into hidden states. \\
        \midrule
        Hidden-state matrix $W_{hh}$ in Eq. \eqref{eqn:rnn} & $N \cdot M\cdot d_h^2$ & External conditions guide the DL model to pass historical information in the hidden states. \\
        \midrule
        Hidden-state bias $\boldsymbol{b}_1$ in Eq. \eqref{eqn:rnn} & $N \cdot M\cdot d_h$ & External conditions guide the DL model to constantly shift the memory in the hidden states. \\
        \midrule
        Input-to-latent matrices $W_{\mu_z x}$, $W_{\sigma_z x}$ in Eq. \eqref{eqn:rnn_encode} & $N \cdot M\cdot d_x \cdot d_z$ & External conditions guide the DL model to encode the current data into the latent space. \\
        \midrule
        Hidden-to-latent matrices $W_{\mu_z h}$, $W_{\sigma_z h}$ in Eq. \eqref{eqn:rnn_encode}, & $N \cdot M\cdot d_h \cdot d_z$ & External conditions guide the DL model to encode past features into the latent space.\\
        \midrule
        Latent vector bias $\boldsymbol{b}_{\mu z}$, $\boldsymbol{b}_{\sigma z}$ in Eq. \eqref{eqn:rnn_encode} & $N \cdot M\cdot d_z$ & External conditions guide the DL model to shift the latent space representations. \\
        \midrule
        Hidden-to-output matrices $W_{xh}$ in Eq. \eqref{eqn:deter_forecast}, $W_{\mu_x h}$, $W_{\sigma_x h}$ in Eqs. \eqref{eqn:prob_forecast1} or \eqref{eqn:rnn_decode} & $N \cdot M\cdot d_h \cdot d_x$ & External conditions guide the DL model to understand past features and forecast the future. \\
        \midrule
        Hidden-to-output bias $\boldsymbol{b}_{2}$ in Eq. \eqref{eqn:deter_forecast}, $\boldsymbol{b}_{\mu x}$, $\boldsymbol{b}_{\sigma x}$ in Eqs. \eqref{eqn:prob_forecast1} or \eqref{eqn:rnn_decode} & $N \cdot M\cdot d_x$ & External conditions guide the DL model shift the forecast data. \\
        \midrule
        Latent-to-output matrix $W_{\mu_x z}$, $W_{\sigma_x z}$ in Eq. \eqref{eqn:rnn_decode} & $N \cdot M\cdot d_z \cdot d_x$ & External conditions guide the DL model decode latent vectors to forecast data. \\
        \bottomrule
    \end{tabular}
    }
    \label{tab:compare_parameter}
\end{table*}

\subsection{Hypernetwork to Make Sequence DL Adaptable}
\label{subsec:hyper}
We aim to enhance the adaptability of DL-based forecasters across varying environments by incorporating external sources of information. A promising and efficient strategy is to employ a Hypernetwork architecture \cite{ha2016hypernetworks}, where one network (the hypernetwork) generates the weights for another (the base network) based on meta-data inputs. By feeding external measurements into the hypernetwork, we can guide the parameter evolution of the base sequence model.

For the $j^{th}$ external source $\{w_{ji}\}_{i=1}^N$, we define a corresponding hypernetwork $g_j(\cdot)$. It is typically implemented as a standard fully connected (FC) layer. At each time step $i$, the output $g_j(w_{ji})$ specifies a subset of parameters for the base RNN $f_{\Theta}(\cdot)$. Thus, we have $g_j(w_{ji})\in \Theta_i$, where we add a subscript $i$ to $\Theta$ to indicate that the parameter set evolves with time and depends on the external meta-knowledge. This brings adaptability to the base RNN. However, as discussed in Section \ref{subsec:dl_forecast}, the parameter set $\Theta_i$ contains multiple components that can be adjusted. Hence, we will balance both computational efficiency and adaptation effectiveness in Section \ref{sec:model} to specify \textbf{meta-representations}, which is sub-parameters in $\Theta_i$ determined by $g_j(w_{ji})$.

\subsection{MoEs for Selecting Informative Meta-Representations}

With increasing data sources in power grid, the number of available external sources is expected to grow substantially. However, leveraging all of them indiscriminately can complicate the forecasting framework and lead to overfitting. To address this, we introduce the Mixture-of-Experts (MoE) architecture. It is a deep learning framework designed to sparsely activate only the most relevant experts, e.g., meta-representations. Compared to traditional selection techniques such as manual feature engineering \cite{nargesian2017learning}, MoE offers the following advantages: (1) It automates the selection process through a gating mechanism, allowing for end-to-end training; (2) It enables dynamic and flexible selection, as the gating outputs can vary over time. 

Mathematically, let $\{g_j(w_{ji})\}_{j=1}^M$ denote a set of $M$ expert networks, where each expert produces a candidate meta-representation. To gate the output for each expert, we introduce another gating network $l(\cdot)$ that understands the past information to determine the current gate values. To designate the input for $l(\cdot)$, we note that, in Eq. \eqref{eqn:rnn}, some parameters will be represented by $g_j(w_{ji})$. Hence, $\boldsymbol{h}_i$ stores past information of loads and external data. Therefore, we can denote $l(\boldsymbol{h}_i)\in \mathbb{R}^M$ to map from $\boldsymbol{h}_i$ to an $M$-dimensional weight vector. The MoE output is then computed as a weighted sum of expert outputs:

\begin{equation}
\label{eqn:moe}
\theta_i = \sum_{j=1}^M l_{ji} \cdot g_j(w_{ji}),
\end{equation}
where $l_{ji} = l(\boldsymbol{h}_i)[j]$ is the gating weight assigned to expert $j$ at time $i$, and $ \sum_{j=1}^M g_{ji} = 1$, where the equation is achieved through the Softmax activation function \cite{zhou2022mixture}. Finally, we enforce that the top-$m$ gating values are activated at each time step to make the selection sparse. This is implemented via
\begin{equation}
\label{eqn:sparse_gate}
l_{ji} = \frac{\exp(\tilde{l}_{ji}) \cdot \mathbb{I}(j \in \mathcal{T}_i)}{\sum_{j \in \mathcal{T}_i} \exp(\tilde{l}_{ji})},
\end{equation}
where $\tilde{l}_{ji}$ is the pre-activation values before the Softmax activation function, and $\mathcal{T}_i \subset \{1, \cdots, M\}$ contains the indices of the top-$m$ experts at time $i$, which are automatically selected by ranking the pre-activation scores $\{\tilde{l}_{ji}\}_{j=1}^M$ and choosing the $m$ highest values. The indicator function \( \mathbb{I}(j \in \mathcal{T}_i) \) ensures that only the top-$m$ experts contribute to the final output. In general, $l(\cdot)$ is made of FC layers with the top-$m$ softmax activation. By Eqs. \eqref{eqn:moe} and \eqref{eqn:sparse_gate}, MoE can generate expressive DL parameter $\theta_i\subset \Theta_i$ as a function of a sparsely weighted sum of meta-representations.

\section{Proposed Model}
\label{sec:model}

In this section, we propose our Meta Mixture of Experts for External data (\text{M\textsuperscript{2}oE\textsuperscript{2}}). The core question is how to specify the DL parameter set $\theta_i\in\Theta_i$ computed from the MoE-hypernetworks in Eq. \eqref{eqn:moe}. Moreover, we provide a worst-case guarantee for this plug-and-play module: with a structural modification, (\text{M\textsuperscript{2}oE\textsuperscript{2}}) plus a base DL model will never perform worse than the DL model. After finalizing our architecture, we demonstrate how to prepare data for the mini-batch-based gradient descent algorithms \cite{hinton2012neural} to train \text{M\textsuperscript{2}oE\textsuperscript{2}}.

\begin{figure*}
    \centering \includegraphics[width=1\linewidth]{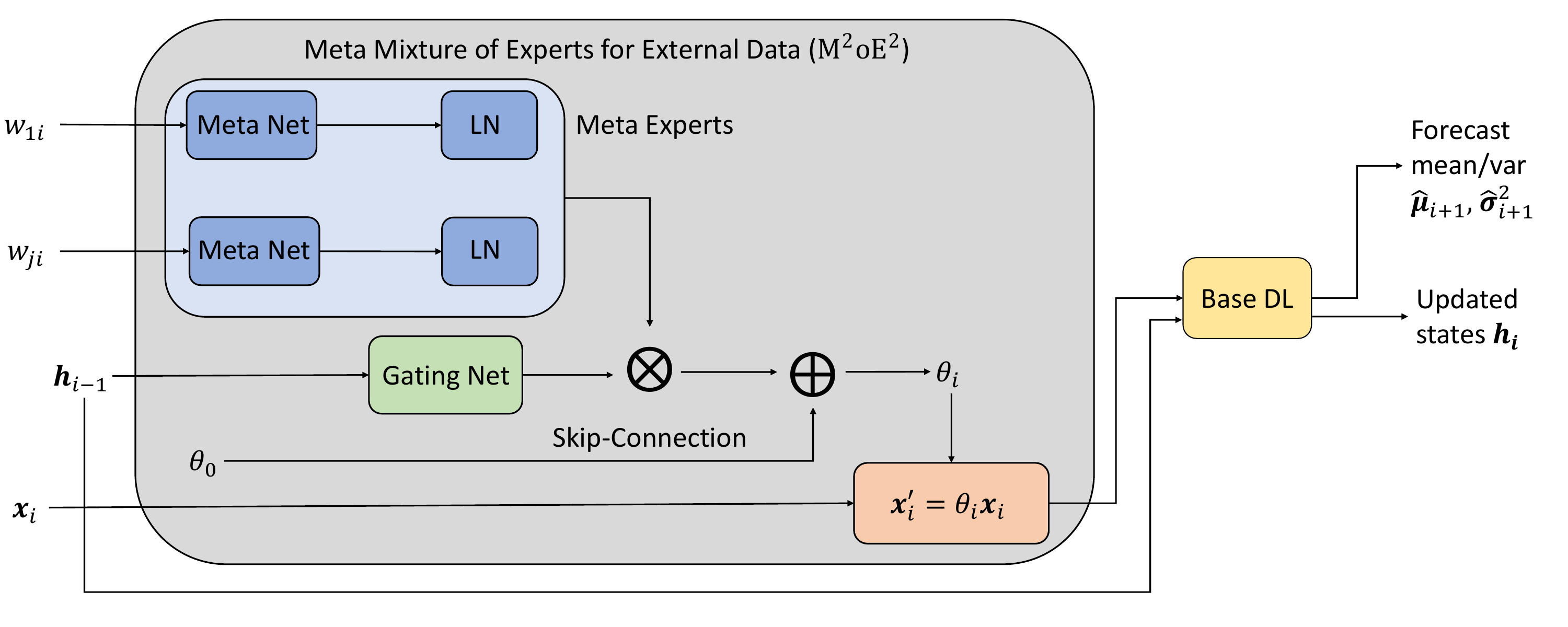}
    \caption{The proposed \text{M\textsuperscript{2}oE\textsuperscript{2}} framework.}
    \label{fig:framework}
    \vspace{-5mm}
\end{figure*}

\subsection{$\theta_i$ Specifications with Computational Efficiency and Adaptation Effectiveness}
\label{subsec:select_thetai}
In the set of candidate components $\Theta_i$ introduced in Section \ref{sec:pre}, both the size and impact of each candidate must be carefully evaluated to determine an appropriate choice for $\theta_i$. Generally, selecting a larger $\theta_i$ incurs higher computational costs, as it requires converting a static parameter into a dynamic, time-varying sequence. On the other hand, choosing a small $\theta_i$ may limit the model’s ability to adapt effectively.
  
To assess these trade-offs, we denote $\boldsymbol{x}_i\in\mathbb{R}^{d_x}$, $\boldsymbol{h}_i\in\mathbb{R}^{d_h}$, and $\boldsymbol{z}\in\mathbb{R}^{d_z}$, where $d_x$, $d_h$, and $d_z$ are the dimensions of $\boldsymbol{x}_i$, $\boldsymbol{h}_i$ and $\boldsymbol{z}_{i+1}$, respectively. We then summarize the size and impact of making each candidate adaptable to external conditions in Table~\ref{tab:compare_parameter}. For example, consider $W_{hx}$ in Eq. \eqref{eqn:rnn}, the training process would involve a $d_x\cdot d_h$ matrix that can be changed according to $M$ external sources at $N$ time slots. This setup allows external conditions to guide how the DL model encodes input load data into hidden states. To select an effective candidate while maintaining manageable computational demands, we make the following observations:

\begin{itemize}
\item \textbf{Generality}: Since we aim to support both deterministic and probabilistic forecasting, parameters associated with the latent variable $\boldsymbol{z}_{i+1}$ are excluded from consideration.
\item \textbf{Efficiency}: In practice, $d_x<d_h$. The input vector $\boldsymbol{x}_i$ typically represents load measurements from a small region (e.g., $d_x\in[1,10]$) since the trained DL models can be applied to different regions. However, the hidden state $\boldsymbol{h}_i$ needs greater capacity to capture historical patterns (e.g., $d_h\in[20,80]]$). To reduce computational load, we exclude $W_{hh}$ from our selection.
\item \textbf{Effectiveness}: As summarized in Table~\ref{tab:compare_parameter}, bias terms generally have less influence compared to weight matrices. Therefore, we recommend making either the input matrix $W_{hx}$ or the output matrix $W_{xh}$ (or their probabilistic variants $W_{\mu_x h}$ and $W_{\sigma_x h}$) adaptable. Since they provide similar empirical performance, we ultimately choose $W_{hx}$ due to its minimal required modifications to existing DL implementations (more details are in Subsection \ref{subsec:archi}).
\end{itemize}

Eventually, we assign the meta-representations in Eq. \eqref{eqn:moe} as $\theta_i = W_{hx}^{(i)}$, enabling a time-varying series of output matrices $\{W_{hx}^{(i)}\}_i$ that dynamically incorporate the load input data.


\subsection{Architecture of \text{M\textsuperscript{2}oE\textsuperscript{2}} with Improvements over Base DL}\label{subsec:archi}

Fig. \ref{fig:framework} presents the architecture of \text{M\textsuperscript{2}oE\textsuperscript{2}}. The overall design modularizes the hypernetwork and the base DL model separately, facilitating straightforward implementation and integration. The model is explained as follows.

First, as described in Eq.~\eqref{eqn:moe} and Eq.~\eqref{eqn:sparse_gate} in Subsection~\ref{subsec:hyper}, the hypernetwork \( g_j \) receives the \( j^{\text{th}} \) external input \( w_{ji} \). Its output is normalized via Layer Normalization (LN), which removes the impact of heterogeneity of the meta data. The blue boxes in Fig. \ref{fig:framework} demonstrate the process.

The produced meta experts are filtered by a gating network  \( l(\boldsymbol{h}_i) \), shown in the green box in Fig. \ref{fig:framework}. Therefore, the gating mechanism can decide what meta data can contribute to the forecasting procedure at the current time. To enhance model robustness and training stability, we introduce a residual (skip) connection. Inspired by the ResNet architecture \cite{he2016deep}, this skip connection ensures that the overall \text{M\textsuperscript{2}oE\textsuperscript{2}} framework will not underperform relative to the base DL model. In other words, the added components can be bypassed if they are not helpful. More specifically, we modify Eq.~\eqref{eqn:moe} as follows:

\begin{equation}
\label{eqn:moe2}
\theta_i = \sum_{j=1}^M l_{ji} \cdot g_j(w_{ji}) + \theta_0,
\end{equation}
where $l_{ji} = l(\boldsymbol{h}_i)[j]$ and \( \theta_0 \) is a static, trainable weight matrix that serves as a shortcut connection. This formulation allows the model to ignore external sources when they are irrelevant. In particular, when all external contributions vanish, i.e., \( \sum_{j=1}^M l_{ji} \cdot g_j(w_{ji}) = 0 \), the meta-representation \( \theta_i \) degrades to \( \theta_0 \), and the \text{M\textsuperscript{2}oE\textsuperscript{2}} behaves identically to the base DL.

Second, we designate the meta-representation \( \theta_i \) as the time-varying input matrix \( W_{hx}^{(i)} \), as discussed in Subsection~\ref{subsec:select_thetai}. However, directly modifying \( W_{hx} \) requires accessing and altering the internal structure of the base DL model, which undermines the goal of a plug-and-play design. To address this, we propose the following alternative formulation:

\begin{equation}
\begin{aligned}
\label{eqn:xprime}
\boldsymbol{x}'_i &= \theta_i \boldsymbol{x}_i.
\end{aligned}
\end{equation}

The left gray box in Fig. \ref{fig:framework} demonstrates the above processes. Then, as shown in the orange and the yellow box, we replace $\boldsymbol{x}_i$ with  $\boldsymbol{x}'_i$ as the input to the base DL model. Hence, the base model can be implemented in either a \textbf{deterministic} form using Eqs.~\eqref{eqn:rnn} and~\eqref{eqn:deter_forecast}, or a \textbf{probabilistic} form based on Eqs.~\eqref{eqn:rnn} and~\eqref{eqn:rnn_encode}--\eqref{eqn:rnn_decode}. The output $\boldsymbol{h}_i$ of the base DL model in turn works as input to the gating neural network in Eq. \eqref{eqn:moe2}. The corresponding training objectives are given by the MSE loss in Eq.~\eqref{eqn:rnn_mse} and the negative ELBO loss in Eq.~\eqref{eqn:rnn_elbo}, respectively. This modular setup allows the base model to be flexibly replaced or extended with any DL architectures.

\vspace{-3mm}

\subsection{Data Preparations to Train \text{M\textsuperscript{2}oE\textsuperscript{2}}}


\subsubsection{Gathering Mini-Batches via Sliding Windows}
We adopt the mini-batch gradient descent strategy \cite{hinton2012neural} to train \text{M\textsuperscript{2}oE\textsuperscript{2}}. While the update rule is standardized, the construction of mini-batches is task-dependent. For instance, we may consider either hourly forecasting (\(K = 1\)) or day-ahead forecasting (\(K = 24\)), where \(K\) is the total steps in forecasting, defined in Subsection~\ref{subsec:dl_forecast}. This setup implies that at each time $i$, the output for each mini-batch should be a 3D tensor of size \( N_{\text{batch}} \times  K \times d_x \), where \( N_{\text{batch}} \) is the number of training instances and \( d_x \) is the load dimension. As for the input, the sequence length should be long enough to capture essential temporal patterns, including trend, seasonality, and short-term momentum. In this paper, the input is a combination of: $(1)$ a fixed-length window spanning the previous week's data and $(2)$ a variable-length window extending from the beginning of the current week up to the time $i$. Then, the model predicts the subsequent \(K\) time steps. In general, we construct each training instance \textbf{by extracting a two-week segment of historical data and appending the following \(K\) time steps as the prediction target}. While each instance has a fixed length, we apply a sliding window to \textbf{extract variable input and fixed output pairs} from the time series, where the input and the output are defined above.

\subsubsection{Standard Gradient-Descent Training}  
Once the training data and loss function are prepared, we apply standard gradient descent to optimize the model parameters over mini-batches. Gradients are computed via backpropagation, and parameters are updated iteratively until convergence. In practice, we implement this process in PyTorch using the commands \texttt{loss.backward()} and \texttt{optimizer.step()}.

\vspace{-3mm}
\section{Experiment}
\label{sec:exp}
\subsection{Settings}
\label{subsec:set}
We evaluate our model and other cutting-edge methods using diverse datasets. Specifically, we have: (1) \textbf{Ashrae Building power consumption} \cite{ref:kaggle_building}. It's a dataset consisting of several buildings' energy consumption and temperature data from the American Society of Heating, Refrigerating and Air-Conditioning Engineers (ASHRAE). (2) \textbf{Spain Loads} \cite{ref:kaggle_spain}. This dataset contains 4 years of electrical consumption, generation, pricing, and weather data for Spain.  (3) \textbf{Tetouan electric power consumption} \cite{ref:kaggle_power}. The data consists of $52,416$ observations of energy consumption on a 10-minute window for Tetouan city located in the north of Morocco. Meteorology data (such as weather, temperature, etc.) is also included.  (4) \textbf{Hourly residential load data in Houston} \cite{ref:kaggle_residential}. The data set contains hourly power usage in kWh from January 2016 to August 2020 in Houston, Texas, USA. A historical weather report is also contained. (5) \textbf{Solar energy} \cite{ref:kaggle_solar}. We also test a renewable energy (i.e., negative loads) dataset, including solar data, temperature, day type, etc. During training, the following external data is employed: air temperature, binary day type indicators (0 for weekday, 1 for weekend), and season labels (0 = Winter, 1 = Spring, 2 = Summer, 3 = Autumn).


To demonstrate the high performance of M\textsuperscript{2}oE\textsuperscript{2}, we introduce the following benchmark methods. (1) \textbf{ARIMA} \cite{hermias2017short}. The classic statistical time-series forecasting method, Autoregressive Integrated Moving Average (ARIMA), is utilized as a benchmark. (2) \textbf{CNN-GRU} \cite{sajjad2020novel}. The model utilizes a GRU to process sequence data and a Convolutional Neural Network (CNN) to extract local features. (3) \textbf{RNN-GRU}. Gated Recurrent Unit (GRU) is coupled with RNN to memorize long-term trends for forecasts \cite{zheng2018short}. (4) \textbf{LSTM} \cite{kong2017short}. Similarly, LSTM employs memory blocks and gates to extract the temporal features.  (5) \textbf{Informer} \cite{zhou2021informer}. Informer is a Transformer-like model that employs the attention mechanism to process time-series data. These methods cover different domains and have diversified architectures. For our \text{M\textsuperscript{2}oE\textsuperscript{2}}, we employ the GRU model. The specific architecture and deployment environment are described in Subsection \ref{subsec:model_arc}.

To evaluate forecasting performance, we consider both deterministic and probabilistic metrics. The \textbf{Mean Squared Error (MSE)} assesses the average squared deviation between predicted means and ground truth values. However, MSE does not capture the uncertainty of predictions. To address this, we incorporate the \textbf{Continuous Ranked Probability Score (CRPS)} \cite{hersbach2000decomposition}, which measures the discrepancy between the predicted cumulative distribution function (CDF) and the empirical CDF of the observed outcome. CRPS provides a proper scoring rule that accounts for both the accuracy and the calibration of probabilistic forecasts. Formally, for a forecast CDF \( F_i \) at time $i$ and an observation \( \boldsymbol{x}_i \), the two metrics are defined as:

\begin{table*}[ht]
\centering
\begin{tabular}{|c|c|c|c|c|c|c|}
\hline
Datasets & \text{M\textsuperscript{2}oE\textsuperscript{2}} & ARIMA & CNN-GRU & RNN-GRU & LSTM & Informer \\
\hline
Ashrae Building Load & {\bf 0.13} & 4.96 & 3.44 & 3.27 & 3.54 & 5.89 \\
\hline
Spain Load & {\bf 0.07} & 6.89 & 1.92 & 2.45 & 2.01 & 2.94 \\
\hline
Tetouan Load & {\bf 0.05} & 0.39 & 0.65 & 0.72 & 0.99 & 2.75 \\
\hline
Houston Residential Load & {\bf 0.04} & 1.41 & 1.67 & 2.18 & 2.38 & 1.69 \\
\hline
Kaggle Solar & {\bf 0.24} & 1.96 & 1.40 & 1.40 & 1.58 & 3.20 \\
\hline
\end{tabular}
\caption{Test MSE ($\times 10^{-2}$) for Different Datasets.}
\label{tab:test_mse}
\end{table*}

\begin{table*}[ht]
\centering
\begin{tabular}{|c|c|c|c|c|c|c|}
\hline
Datasets & \text{M\textsuperscript{2}oE\textsuperscript{2}} & ARIMA & CNN-GRU & RNN-GRU & LSTM & Informer \\
\hline
Ashrae Building Load & {\bf 1.90} & 13.01 & 11.44 & 11.26 & 11.63 & 14.49 \\
\hline
Spain Load & {\bf 1.29} & 24.64 & 7.49 & 9.09 & 7.76 & 10.19 \\
\hline
Tetouan Load & {\bf 0.89} & 3.59 & 4.84 & 4.97 & 5.71 & 11.44 \\
\hline
Houston Residential Load & {\bf 0.80} & 6.38 & 6.41 & 7.59 & 7.85 & 6.25 \\
\hline
Kaggle Solar & {\bf 2.03} & 7.11 & 4.24 & 4.28 & 4.56 & 7.25 \\
\hline
\end{tabular}
\caption{Test CRPS ($\times 10^{-2}$) for Different Datasets.}
\label{tab:test_crps}
\end{table*}

\begin{equation}
\begin{aligned}
\text{MSE} &= \frac{1}{N} \sum_{i=1}^N \left( \hat{\boldsymbol{\mu}}_i - \boldsymbol{x}_i \right)^2,\\
\text{CRPS} &= \frac{1}{N}\sum_i^N\text{CRPS}_i = \frac{1}{N} \int_{-\infty}^{\infty} \left( F_i(\boldsymbol{z}) - \mathbf{1}\{\boldsymbol{z} \geq \boldsymbol{x}_i\} \right)^2 d\boldsymbol{z},
\end{aligned}
\end{equation}
where \( \hat{\boldsymbol{\mu}}_i \) is the predicted mean and \( \mathbf{1}\{\boldsymbol{z} \geq \boldsymbol{x}_i\} \) is the Heaviside step function representing the observed outcome. In our model, we assume that the predictive distribution of each $1$-dimensional forecast entry has the Gaussian mean \( \hat{\mu}_i \) and standard deviation \( \hat{\sigma}_i \). This allows us to compute the CRPS value at time $i$ in a closed form as:

\begin{equation}
\begin{aligned}
\text{CRPS}_i &=  \hat{\sigma}_i \Bigg( \frac{1}{\sqrt{\pi}} - 2\phi\left( \frac{x_i - \hat{\mu}_i}{\hat{\sigma}_i} \right) \\
&- \frac{x_i - \hat{\mu}_i}{\hat{\sigma}_i} \left(2\Phi\left( \frac{x_i - \hat{\mu}_i}{\hat{\sigma}_i} \right) - 1 \right) \Bigg),
\end{aligned}
\end{equation}
where \( \phi(\cdot) \) and \( \Phi(\cdot) \) denote the probability density function and cumulative distribution function of the standard normal distribution, respectively. For the multi-dimensional case, we can compute the average CRPS across all entries in $\boldsymbol{x}_i$.

\subsection{Model Architectures and Deployment Environments}
\label{subsec:model_arc}

The model is implemented in PyTorch and trained on a device with an Apple M2 chip (8-core CPU) and 16 GB of memory, using Python 3.9. Training was conducted using the Adam optimizer (learning rate = $1\times 10^{-3}$) for $300$ epochs with a batch size of $16$.

In \text{M\textsuperscript{2}oE\textsuperscript{2}}, the base DL model is a $4$-layer GRU network. The encoder of the base DL (see Eq. \eqref{eqn:rnn_encode}) takes in the $\boldsymbol{x}'_i$ with the dimension $d_{x'}=40$. Then, the temporal information is stored in the hidden state $\boldsymbol{h}_i$ with the dimension $d_h= 64$. The decoder of the base DL (see Eq. \ref{eqn:rnn_decode}) converts from the hidden state $\boldsymbol{h}_i$ to the random latent vector $\boldsymbol{z}_i$ with the dimension $d_z=32$. Finally, a fully-connected linear layer is utilized to forecast the future $K=3$ steps of data using $\boldsymbol{z}_i$.

To produce $\boldsymbol{x}'_i$, we construct the meta representation and use Eq. \eqref{eqn:xprime} to map from $\boldsymbol{x}_i$ to $\boldsymbol{x}'_i$. We consider univariate forecasting so that $d_x=1$. Then, we utilize $2$-layer meta networks with tanh activation functions and $d_x\times d_{x'} = 40$ hidden units. The output of the meta-network ($g_j(\cdot)$) is reshaped to a $d_x\times d_{x'}$ matrix $g_j(w_{ji})$. Totally, we have $M=3$ experts ($\theta_i$) from the temperature, day-type, and season-type data. We select the top-$2$ experts by ranking the gating scores in MoE. The weighted sum of $g_j(w_{ji})$ is combined with a static parameter set $\theta_0\in \mathbb{R}^{1\times 40}$ in Eq. \eqref{eqn:moe2} to produce $\theta_i\in\mathbb{R}^{1\times 40}$. By now, the complete architecture in Fig. \ref{fig:framework} is fully introduced. Finally, we have a KL divergence term with the weight $\lambda=0.01$, as described in Eq. \eqref{eqn:rnn_elbo}.

\subsection{General Results}

For fair comparisons, we test all methods with the same load, renewable energy, and external data, explained in Section \ref{subsec:set}. However, \text{M\textsuperscript{2}oE\textsuperscript{2}} treats external factors as meta-knowledge, but others treat them as the direct input. Tables \ref{tab:test_mse} and \ref{tab:test_crps} present the average test MSE and CRPS for different methods. In general,  \text{M\textsuperscript{2}oE\textsuperscript{2}}'s test error is around $5\%\sim 50\%$ of the error in other methods, demonstrating the high capacity and generalizability of \text{M\textsuperscript{2}oE\textsuperscript{2}} across different data sources. 

The excellent performance relies on the following designs. First, by treating external data as meta-knowledge, the hypernetwork in \text{M\textsuperscript{2}oE\textsuperscript{2}} essentially creates a series of DL models that efficiently adapt to the environment information. Specifically, the temperature values and date-type labels are relatively flat (see Fig. \ref{fig:one_week_data}) compared to the load data. Hence, other methods may view them as constant. However, our meta-representation always keeps the context information and avoids naively disregarding their contributions. For example, a small change in the external value in $w_{ji}$ can cause a big change in the parameter set $\theta_i$. Second, \text{M\textsuperscript{2}oE\textsuperscript{2}} contains an MoE structure to filter the multiple external sources and select the most useful one. Others include all of them and easily face overfitting. Third, the skip-connection in \text{M\textsuperscript{2}oE\textsuperscript{2}} guarantees that the external information will never deteriorate performance.  

\begin{figure*}
    \centering \includegraphics[width=1\linewidth]{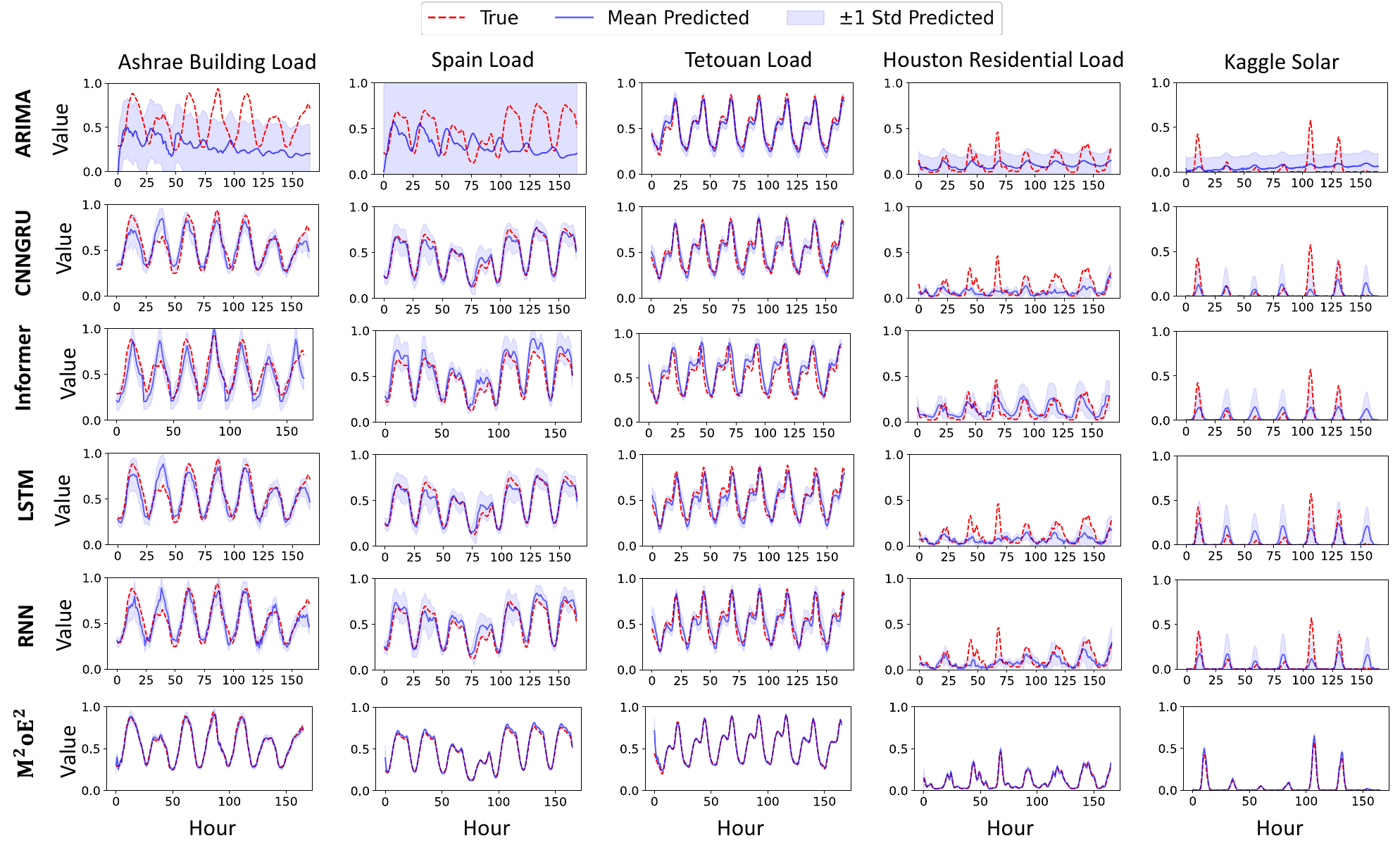}
    \caption{The forecast and the true loads in week $1$.}
    \label{fig:All_results_plot1}
\end{figure*}

\begin{figure*}
    \centering \includegraphics[width=1\linewidth]{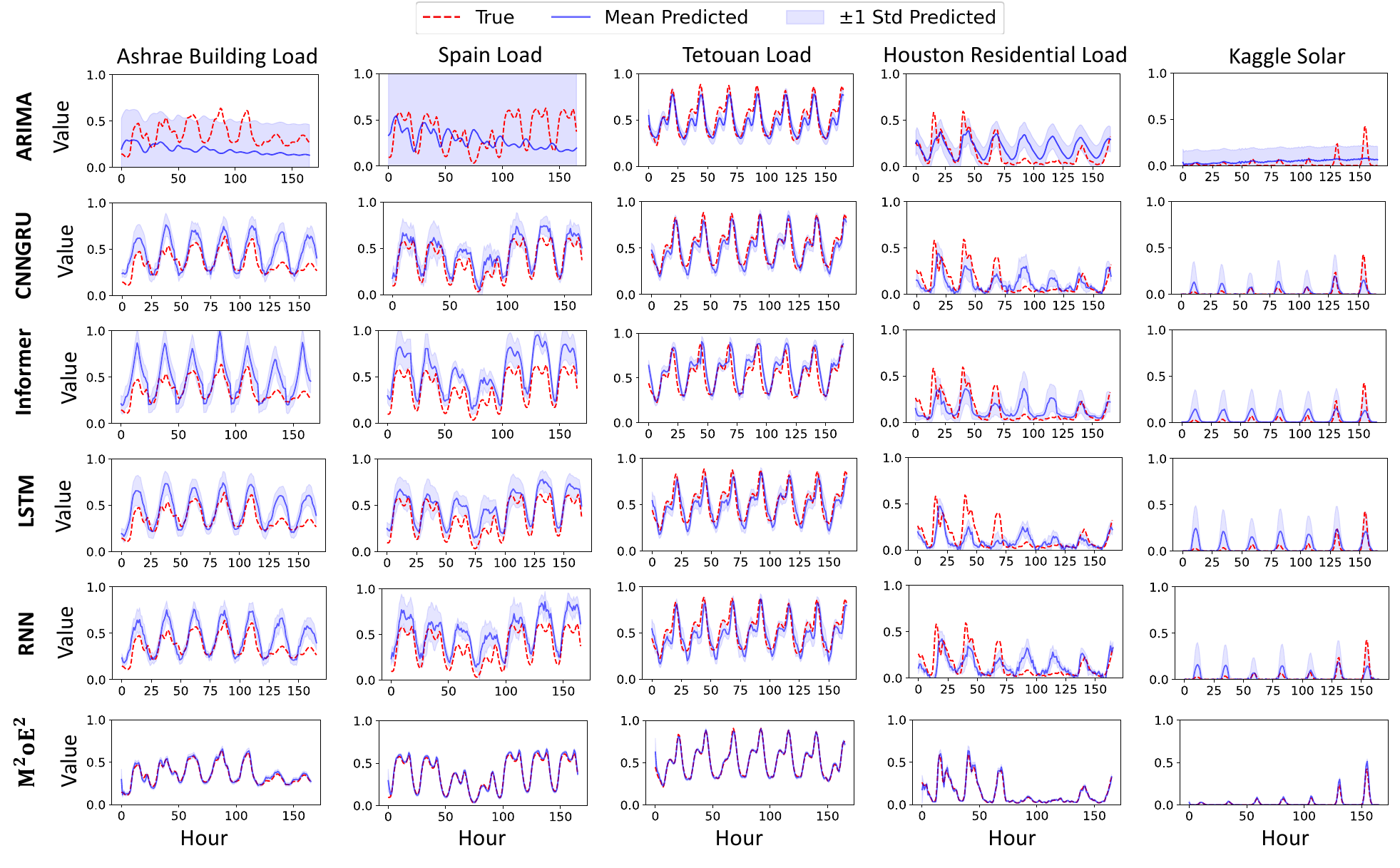}
    \caption{The forecast and the true loads in week $3$.}
    \label{fig:All_results_plot2}
\end{figure*}

\subsection{Visualizations for Forecast Hourly Loads}

To directly understand how good the performance is, we visualize the true data and the forecast hourly loads for two random weeks in Fig. \ref{fig:All_results_plot1} and \ref{fig:All_results_plot2}, respectively. In general, we have the following critical observations. First, \text{M\textsuperscript{2}oE\textsuperscript{2}} not only has the best mean forecast fitting but also has a very small variance prediction, which implies that the model is confident in its prediction. This is extremely beneficial for the downstream operations and optimization. Second, the only high variance for \text{M\textsuperscript{2}oE\textsuperscript{2}} appears in the first few hours' forecasting. This is because the model has no or limited new observations in the first few hours. However, after acquiring several new data points, \text{M\textsuperscript{2}oE\textsuperscript{2}} quickly produces confident forecasts for the subsequent hours. In contrast, other methods don't have such converged patterns of the variance prediction. Third, there are non-stationary load curves, e.g., the Ashrae building data in Fig. \ref{fig:All_results_plot2}, the weekend load patterns are distinct from those of weekdays. Under this setting, \text{M\textsuperscript{2}oE\textsuperscript{2}} is the only method that can adapt to the day types and change its forecasting patterns.

\section{Conclusion}
\label{sec:con}
In this paper, we propose \text{M\textsuperscript{2}oE\textsuperscript{2}} to efficiently utilize environmental and power system data for load forecasting. Our proposed model is highly accurate, efficient, and robust across diverse datasets. \text{M\textsuperscript{2}oE\textsuperscript{2}}'s overall error is only $5\%\sim 50\%$ of the error from the cutting-edge statistical or DL methods. Such an excellent performance arises from $(1)$ the high capacity and adaptation power due to meta-representations, $(2)$ the efficient and robust processing of external sources by using MoE structures, and $(3)$ the guaranteed improvement over a base DL model, thanks to the residual architecture. Consequently, \text{M\textsuperscript{2}oE\textsuperscript{2}} provides a novel framework to incorporate external information while keeping improving with the state-of-the-art DL models.

\clearpage
\bibliographystyle{IEEEtran}
\bibliography{IEEEabrv,reference}
\end{document}